\documentclass[acmtog]{acmart}
\acmSubmissionID{237}

\usepackage{xr}
\usepackage[ruled]{algorithm2e} %
\usepackage{graphicx}
\usepackage{amsmath}
\usepackage{booktabs}
\usepackage{paralist}

\usepackage{xcolor}
\definecolor{citecolor}{RGB}{30,102,235}

\usepackage[capitalize]{cleveref}

\usepackage{soul}

\usepackage{subcaption}
\usepackage{lipsum}  
\usepackage{pifont}
\usepackage{enumitem}
\usepackage{physics}

\crefname{section}{Sec.}{Secs.}
\Crefname{section}{Section}{Sections}
\Crefname{table}{Table}{Tables}
\crefname{table}{Tab.}{Tabs.}
\crefname{subsection}{subsection}{subsections}

\DeclareMathOperator*{\argmax}{arg\,max}

\makeatletter
\DeclareRobustCommand\onedot{\futurelet\@let@token\@onedot}
\def\@onedot{\ifx\@let@token.\else.\null\fi\xspace}

\def\eg{\emph{e.g}\onedot} 
\def\ie{\emph{i.e}\onedot} \def\Ie{\emph{I.e}\onedot}

\definecolor{mydarkblue}{rgb}{0,0.08,1}
\definecolor{mydarkgreen}{rgb}{0.02,0.6,0.02}
\definecolor{mydarkorange}{rgb}{0.40,0.2,0.02}
\definecolor{mypurple}{RGB}{111,0,255}
 \definecolor{mycyan}{rgb}{0.02,0.6,0.42}
\definecolor{mygold}{rgb}{0.75,0.6,0.12}

\newif\ifdraft
\drafttrue

\ifdraft
    \newcommand{\aec}[1]{{\color{blue}\textbf{AE:} #1}}
    \newcommand{\dcc}[1]{{\color{red}\textbf{Danny:} #1}}
    \newcommand{\ync}[1]{{\color{mydarkgreen}\textbf{YN:} #1}}
    \newcommand{\opc}[1]{{\color{orange}[\textbf{OP:} #1]}}

    \newcommand{\work}[1]{{\color{mygold}#1}}

\else
    \newcommand{\aec}[1]{}
    \newcommand{\dcc}[1]{}
    \newcommand{\ync}[1]{}
    \newcommand{\opc}[1]{}

    \newcommand{\work}[1]{{\color{black}#1}}
    
\fi

\newcommand{\myparagraph}[1]{\vspace{0pt}\paragraph{#1}}

\newcommand{\vect}[1]{\boldsymbol{\mathbf{#1}}}

\newcommand{\notation}[1]{\ensuremath{#1}\xspace}

\newcommand{\w}{\notation{\mathcal{W}}}

\newcommand{\wplus}{$\mathcal{W}^+$\xspace}

\newcommand{\p}{\notation{\mathcal{P}}}
\newcommand{\pbeta}{\notation{\mathcal{P}_{\beta}}}
\newcommand{\pplus}{\notation{\mathcal{P}^+}}

\newlist{todolist}{itemize}{2}
\setlist[todolist]{label=$\square$}

\makeatletter
\newcommand*{\addFileDependency}[1]{%
  \typeout{(#1)}
  \@addtofilelist{#1}
  \IfFileExists{#1}{}{\typeout{No file #1.}}
}
\makeatother

\newcommand*{\myexternaldocument}[1]{%
    \externaldocument{#1}%
    \addFileDependency{#1.tex}%
    \addFileDependency{#1.aux}%
}

\myexternaldocument{supp}

\SetAlFnt{\small}
\SetAlCapFnt{\small}
\SetAlCapNameFnt{\small}
\SetAlCapHSkip{0pt}

\acmJournal{TOG}

\begin{document}
\title{Facial Reenactment Through a Personalized Generator}

\author{Ariel Elazary}
\affiliation{
    \institution{Tel-Aviv University}
    \country{Israel}
}
\author{Yotam Nitzan}
\affiliation{
    \institution{Tel-Aviv University}
    \country{Israel}
}
\author{Daniel Cohen-Or}
\affiliation{
    \institution{Tel-Aviv University}
    \country{Israel}
}

\begin{teaserfigure}
    \centering
    \vspace{-2mm}
    \includegraphics[width=0.97\linewidth]{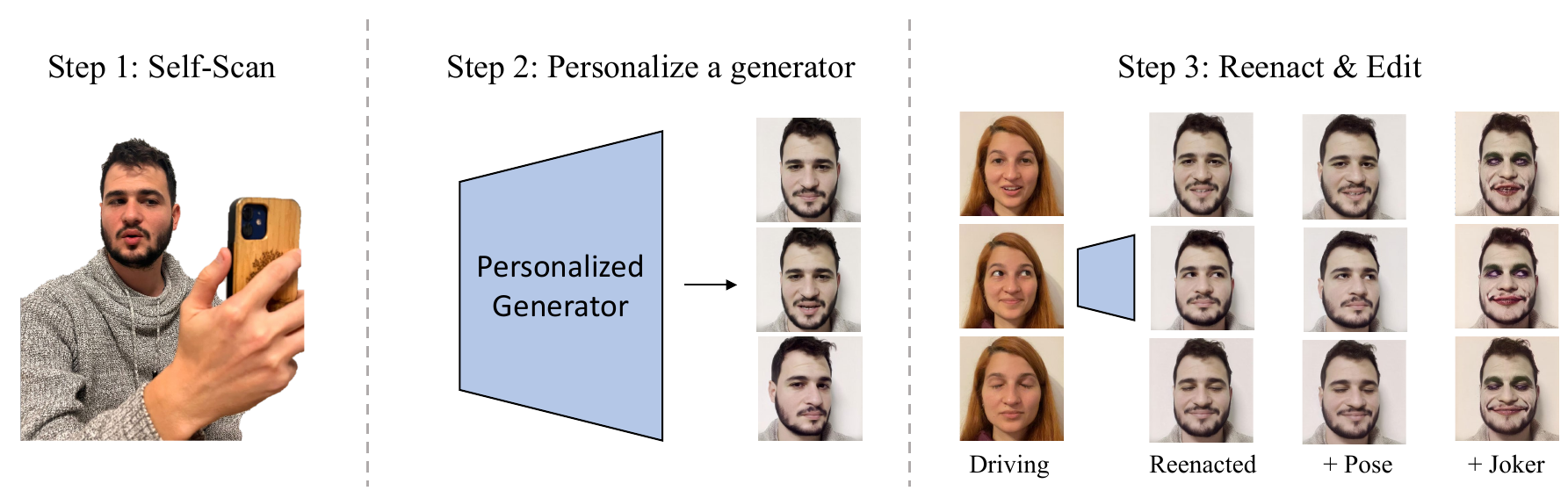}
    \caption{
        We present a novel approach to reenact an individual using only a casual \emph{self-scan}.
        Using a captured self-scan (Step 1), we train a personalized image generator, tailored specifically to that individual (Step 2).
        The personalized generator guarantees the preservation of the individual's appearance and facilitates video-driven reenactment. 
        Thanks to the generator's semantic latent space, the generated video can be edited and stylized in post-process (Step 3).
        }
    \label{fig:teaser}
\end{teaserfigure}

\begin{abstract}

In recent years, the role of image generative models in facial reenactment has been steadily increasing.
Such models are usually subject-agnostic and trained on domain-wide datasets.
The appearance of the reenacted individual is learned from a single image, and hence, the entire breadth of the individual's appearance is not entirely captured, leading these methods to resort to unfaithful hallucination.
Thanks to recent advancements, it is now possible to train a personalized generative model tailored specifically to a given individual.
In this paper, we propose a novel method for facial reenactment using a personalized generator.
We train the generator using frames from a short, yet varied, self-scan video captured using a simple commodity camera.
Images synthesized by the personalized generator are guaranteed to preserve identity.
The premise of our work is that the task of reenactment is thus reduced to accurately mimicking head poses and expressions.
To this end, we locate the desired frames in the latent space of the personalized generator using carefully designed latent optimization.
Through extensive evaluation, we demonstrate state-of-the-art performance for facial reenactment.
Furthermore, we show that since our reenactment takes place in a semantic latent space, it can be semantically edited and stylized in post-processing.
\end{abstract}



\maketitle
\balance
\section{Introduction}

Facial reenactment is the task of synthesizing a realistic portrait video of a target person that replicates the actions of a person appearing in a driving video.
The ability to do so is essential for movie postproduction, augmented reality (AR), virtual reality (VR) and telepresence applications.
The main challenge of this task is faithfully representing the identity, or appearance, of the target person, in varying conditions.

For years, works approaching this task took a graphics perspective, explicitly modeling 3$D$ geometry, usually relying on morphable mesh models~\cite{kim2018deep,zheng2022avatar,ichim2015dynamic,thies2016face2face,hu2017avatar,thies2019deferred,grassal2022neural,lombardi2018deep}.
Recently, several works have leveraged $2D$ image generative models for the task~\cite{lin20223d,wang2021one,sun2022next3d,abdal2022video2stylegan,zakharov2019few}.
These methods offer superior image quality than those based on traditional $3D$ graphics as they inherently support more facial components including hair, the mouth cavity, and accessories like glasses.
Moreover, a notable advantage of using generative models is that they support intuitive semantic control, such as aging or adding glasses, through their latent spaces~\cite{patashnik2021styleclip,shen2020interfacegan,abdal2020styleflow} and out-of-domain stylization~\cite{gal2021stylegan,zhu2021mind,ojha2021few} (see~\cref{fig:teaser}).

Although methods relying on image generative models have many advantages, they compromise identity preservation.
Previous works have primarily used one or just a few images (\eg, $8$) to learn the appearance of the target person~\cite{lin20223d,sun2022next3d,wang2021one,abdal2022video2stylegan}. 
Using one or very few images is usually claimed as an advantage as they are easy to obtain.
However, the full variance and aspects of a person's appearance clearly cannot be learned in this setting. 
To complete missing information, such methods resort to hallucination based on generic face priors learned from many different individuals.
Realistically, beyond data acquisition, another factor popularizing this setting was the lack of methods to train a generator to preserve identity as portrayed in a reasonably sized personal photo album.

In this work, we leverage recent advancements in generative modelling and employ an \emph{image generator} that is \emph{personalized} -- \ie trained specifically for a given person.
As input, we take a \emph{self-scan} -- a short RGB video that captures the target person in a variety of viewpoints and facial expressions.
Such a self-scan video, albeit short, is sufficient to personalize a domain-wide (\ie, faces) image generator \cite{karras2020analyzing} and enables a faithful modelling of the person's identity.
Importantly, the  self-scan is still easy to obtain as it is casually captured using any standard camera and does not require a high-cost apparatus nor a controlled environment nor a strict script to follow.

To personalize the generative model, we build upon MyStyle~\cite{nitzan2022mystyle}.
Requiring roughly $100-300$ images of an individual, MyStyle fine-tunes a generic face generator~\cite{karras2020analyzing} to model the individual's identity.
Importantly, other properties of the generator, such as the image quality and semantic latent space are preserved. 
Using the personalized generator immediately mitigates the challenge of preserving identity.
Reenactment is then reduced to locating a code in the generator's latent space that imitates the head pose and expression of the person in a driving frame, which we perform using \textit{latent optimization}.

This optimization is nevertheless challenging in itself. 
Common representations for pose and expression are often entangled with the identity information.
For example, facial keypoints~\cite{kartynnik2019real} are often used to track facial motion and specifically used to indicate on the person's pose and expression.
However, keypoints also indicate the facial geometry and were shown to encode identity~\cite{nixon1985eye}.
Due to this fact, previous works leveraging keypoints were limited to only have a person reenacting themselves (\ie, \textit{self-reenactment})~\cite{zakharov2019few}.
We overcome this difficulty by using a keypoint-driven representation from which we eliminate most identity-related information.
Concisely, we use a carefully selected subset of keypoints and represent each with a relative location, with respect only to other keypoints belonging to the same facial part (\eg, left eye).
To capture additional information that is overlooked by the keypoints, such as the presence of teeth, we directly use the driving frame's pixels in specific regions.
We optimize the latent code to according to these two terms.
Note that neither term is perfectly agnostic to identity, which might hinder identity preservation if used in a different setting.
However, the personalized generator is constrained to identity-preserving results and the  optimization can only recover the nearest possible solution.

A key challenge in video generation is maintaining temporal consistency.
Although two consecutive frames are similar, two separate optimization processes might converge to latent codes that are unnecessarily different. 
To mitigate temporal inconsistency, we initialize each optimization process with a slightly noised version of the latent code produced for the previous frame.
Since optimization is highly susceptible to its initialization, this approach biases the optimization to converge to a nearby latent and hence, minimizes unnecessary perturbation between the two consecutive frames.

Through quantitative and qualitative evaluation, we demonstrate that our method is highly effective at facial reenactment.
Specifically, we find that our method excels at identity preservation, surpassing the performance of recent state-of-the-art methods, and performs at least on par in terms of transferring pose and expression.

\section{Related Work}

\subsection{Facial Reenactment}
Facial Reenactment was traditionally approached from a $3$D-graphics perspective.
Such methods usually start by collecting significant footage of the target face and reconstructing an explicit $3$D mesh model, using $3$D Morphable Models~\cite{blanz1999morphable}.
At test time, reenactment is performed by transferring pose and expression from the $3$D model of the driving video to that of the target 3D face model~\cite{thies2016face2face,thies2015real,hu2017avatar,ichim2015dynamic,garrido2014automatic,cao20133d,cao2014displaced,bouaziz2013online,blanz2003reanimating, grassal2022neural}.
A thorough introduction of facial reconstruction and $3$DMM is provided in recent surveys~\cite{zollhofer2018state,egger20203d}.
Explicitly modeling faces in $3$D is challenging and most methods struggle with modelling hair, mouth cavity and accessories like glasses.

Recently, different approaches were proposed to sidestep these challenges.
First, several works avoided $3$D modelling altogether by warping an image of the target face~\cite{averbuch2017bringing,siarohin2019first,drobyshev2022megaportraits}.
With this approach, faithful changes to pose and expression possible are relatively limited, often leading to undesired artifacts.
A second approach models faces using other $3$D representations such as neural implicit representations~\cite{gafni2021dynamic,zheng2022avatar} and point clouds~\cite{Zheng2022pointavatar}.
Last, and most relevant to our work, several recent works have incorporated $2$D generative models, mostly GANs~\cite{goodfellow2014generative}, to their pipelines.
Some of these works have proposed pipelines that are entirely identity-agnostic.
\Ie, a generative model is trained on a domain-wide dataset, and a single image of the individual is used as a starting point for manipulation~\cite{sun2022next3d,lin20223d,wang2022latent,wang2021one,abdal2022video2stylegan,hong2022depth}.

A single image is insufficient to learn the entire breadth of appearances of an individual, leading such methods to resort to unfaithful hallucination.
Accordingly, other works have incorporated more footage of the individual on top of generic knowledge of faces.
\citet{kim2018deep} and the concurrent work by \citet{wang2023styleavatar} classically perform reenactment using $3$DMM but employ person-specific generative models to transform $3$DMM-based renders to realistic images. 
Inevitably, they still partially suffer from limitations stemming from using $3$D mesh models, such as not modeling the mouth cavity.
A different approach, also taken by this work, is pretraining the model on a domain-wide dataset and then fine-tuning it for a specific individual.
\citet{zakharov2019few} pretrain on VoxCeleb~\cite{Chung18b}, a low-resolution ($\leq 256$p) video dataset, limiting the eventual quality of their output. 
Their method is also limited to only support an individual reenacting themselves.
\citet{cao2022authentic} pretrain on an unpublished dataset that is laborious to collect, featuring hundreds of identities, performing scripted facial expressions while being captured by 90 cameras simultaneously. 
In contrast, our domain-wide model is an off-the-shelf StyleGAN2~\cite{karras2020analyzing} model, trained on FFHQ~\cite{karras2019style} -- a high-resolution ($1024$p) image dataset, that is crawled off the Internet.
Our simple pretraining approach is significantly more accessible, does not compromise output quality, and could naturally be improved by using future image datasets.

\subsection{Generative Prior}

In recent years, generative models have immensely improved in countless foundational tasks such as unconditional synthesis~\cite{karras2019style,brock2018large}, image-to-image translation~\cite{isola2017image,zhu2017unpaired} and text-condition synthesis~\cite{rombach2022high,ramesh2022hierarchical}.
Noticeably, unconditional GANs~\cite{goodfellow2014generative}, and specifically StyleGAN~\cite{karras2019style,karras2020analyzing}, have drawn attention for the meaningful representations that emerge in the generator's latent space.
Such meaningful representations make the generator useful for numerous downstream applications, beyond simply image synthesis.
One line of works find semantic controls in the generator's latent space which enable controlled synthesis and image editing pipelines~\cite{patashnik2021styleclip,abdal2020styleflow,voynov2020unsupervised,shen2020closedform,shen2020interfacegan,harkonen2020ganspace}.
Other works leverage these pretrained generators as a prior for downstream tasks ranging from image enhancement~\cite{menon2020pulse, yang2021gan, chan2021glean, richardson2021encoding,wang2021towards,luo2020time,pan2021exploiting, gu2020image}, facial reenactment~\cite{nitzan2020dis,lin20223d,sun2022next3d,abdal2022video2stylegan} and even discriminative tasks, such as regression~\cite{xu2021generative}.

Methods leveraging unconditional GANs often start with locating a latent code that corresponds to a desired image.
This process, often known as \emph{projection}, can be carried out using a trained encoder~\cite{richardson2021encoding,nitzan2020dis,tov2021designing,xu2021generative,pidhorskyi2020adversarial} or using optimization~\cite{abdal2019image2stylegan,parmar2022spatially,zhu2020domain,gu2020image,karras2020analyzing}.
Encoding is significantly faster than optimization, but is also less accurate.

Regardless of the projection method, the desired image might not exist within the output space of the generator, greatly limiting the applicability of this approach.
To overcome this challenge, several works have proposed to slightly fine-tune the generator in order to improve its expressiveness on specific inputs~\cite{pan2021exploiting,roich2021pivotal,nitzan2022mystyle,bau2020semantic}.
Within this line of works,~\citet{nitzan2022mystyle} recently proposed MyStyle -- a method to fine-tune the generator to faithfully model a specific person.

In this work, we rely on MyStyle's personalized generator as prior for reenactment. 
Technically, we locate the desired frame by projecting aspects of the driving frame such as facial keypoints and the eyes and mouth regions.

\section{Method}

We aim to generate a portrait video in which a target person, appearing in a self-scan video, is repeating the actions occurring in a driving video.
The self-scan is a short, yet varied, RGB video used to learn the appearance of the target individual.
The simple process of capturing a self-scan is discussed in~\cref{subsec:capture}.
Our approach is composed of two steps. 
First, using the self-scan, we train a personalized image generator, building upon MyStyle~\cite{nitzan2022mystyle} (\cref{subsec:training}).
The personalized generator is a universal representation of the individual, and can be used in conjunction to any driving video.
Second, we perform latent optimization to locate a latent code of the personalized generator that faithfully represents the actions in a driving frame (\cref{subsec:optimization}).

\subsection{Self-Scan Capture}
\label{subsec:capture}

To authentically model an individual's appearance, one must observe them in a wide variety of settings.
For example, the best, or perhaps only, way of knowing how a person looks like when smiling is to actually see them smiling beforehand.

We find that a short, single-take, RGB video can include sufficiently varied head poses and expressions. 
We call such videos \emph{self-scans}, as they can be conveniently captured by the individual themselves.
Note that the individual does not need to be following any script.
Instead, the self-scan should intuitively include the poses and expressions that one desires to be possible for the generated video.
One does not need to exhaust all combinations of poses and expressions, and performing each once is sufficient.
As we demonstrate later, our method enables disentangling and recomposing poses and expressions that were not seen jointly in the self-scan.

In practice, for the experiments in this paper, we instructed several individuals to capture a $20\text{-}30$ seconds, portrait video of themselves with varying head poses and expressions that could be used in a social interaction, like conversation.

\subsection{From a Self-Scan to a Personalized Generator}
\label{subsec:training}

At the core of our method is an unconditional personalized generator $G$, mapping a latent code to an image.
To train it, we employ the recently proposed method MyStyle~\cite{nitzan2022mystyle}.
Using an image dataset of an individual, MyStyle slightly fine-tunes StyleGAN2~\cite{karras2020analyzing}, a generic face image generator, to obtain a generator that faithfully preserves individual's appearance.

\myparagraph{Creating the Training Set.}
StyleGAN2 and MyStyle are trained on aligned portrait images.
We therefore start with \emph{smoothly} aligning and cropping all frames in the self-scan~\cite{tzaban2022stitch}.

MyStyle was demonstrated effective when used with a dataset of $100\text{-}300$ randomly curated images.
A self-scan, however, is composed of significantly more frames, often over a thousand.
Increasing the training set size slows training and inference times, due to increased dimensionality~\cite{nitzan2022mystyle}. 
We observe that not all frames are required for training.
Many frames are similar to each other -- consecutive frames, for example, are nearly duplicates.
Using two frames that are interchangeable adds negligible diversity and information to the generator.
Intuitively, one of these frames could be omitted.

Let us denote the set of aligned self-scan crops by $S$.
We wish to find a subset $T \subset S$, where $T$ contains a predetermined number of frames, marked $N$, and maximizes the subset's ``diversity''.
Following previous works~\cite{ojha2021few,nitzan2022mystyle}, we use the average pairwise LPIPS~\cite{zhang2018unreasonable} distances as a measure of a set's diversity.
Identifying the optimal subset $T$ is known to be NP-hard but a greedy algorithm was proven to be quite effective~\cite{ravi1994heuristic}.
The greedy algorithm iteratively adds the element of the set $S \setminus T$ whose minimal distance to any element of current subset $T$ is greatest.
See pseudo-code description in~\cref{alg:subset} in the supplemental.
We find that $N=250$ is sufficient.

\myparagraph{MyStyle training.}

\begin{figure}
    \centering
     \includegraphics[width=\linewidth]{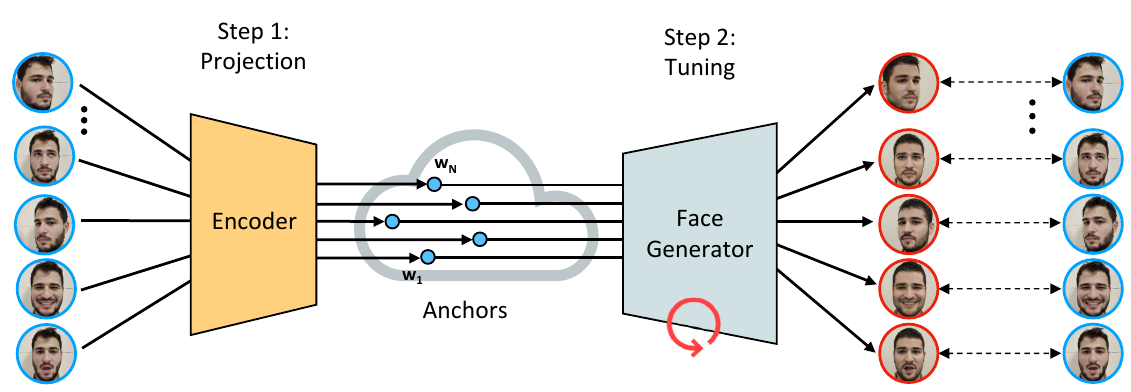}
    \caption{
    An illustration of MyStyle's tuning method.
    A set of $N$ portrait images of the target individual are projected into StyleGAN’s latent space.
    The result is a set of anchors that are the nearest possible neighbors.
    We then tune the generator to reconstruct the input images from their corresponding anchors according to~\cref{eq:mystyle_training}.
    }
    \label{fig:tuning}
\end{figure}

Using the filtered self-scan we next train MyStyle \cite{nitzan2022mystyle}.
Due to MyStyle's central role in training and later on in reenactment, we concisely repeat the training details.

We start with our training set $T = \{t_i\}_{i=1}^N$ and a StyleGAN2 generator~\cite{karras2020analyzing} pretrained on FFHQ~\cite{karras2019style}.
Each image $t_i$ is first inverted into StyleGAN's latent space \w, yielding a latent code $w_i$, called \textit{anchor}. 
Inversion is performed using an off-the-shelf inversion encoder, whose architecture follows~\citet{richardson2021encoding}.
Now, the pretrained StyleGAN2 generator is fine-tuned using pivotal tuning~\cite{roich2021pivotal} -- \ie, to better reconstruct the training images from their corresponding anchors.
Formally, the training loss is given by
\begin{equation}
\label{eq:mystyle_training}
\begin{split}
    \mathcal{L}_{\text{mystyle}} = \sum_{i=1}^N \; [ \mathcal{L}_{\text{lpips}}(G(w_i), t_i) +
    \lambda_{\text{pixel}}\norm{G(w_i) - t_i}_2],
\end{split}    
\end{equation}
where $\mathcal{L}_{\text{lpips}}$ is the LPIPS loss~\cite{zhang2018unreasonable} and $\lambda_{\text{pixel}}=10$ is a hyperparameter balancing the two terms.

We illustrate the training process in~\cref{fig:tuning}.

\subsection{Reenactment by Latent Optimization}
\label{subsec:optimization}

\begin{figure}[h]
    \centering
    \includegraphics[width=\linewidth]{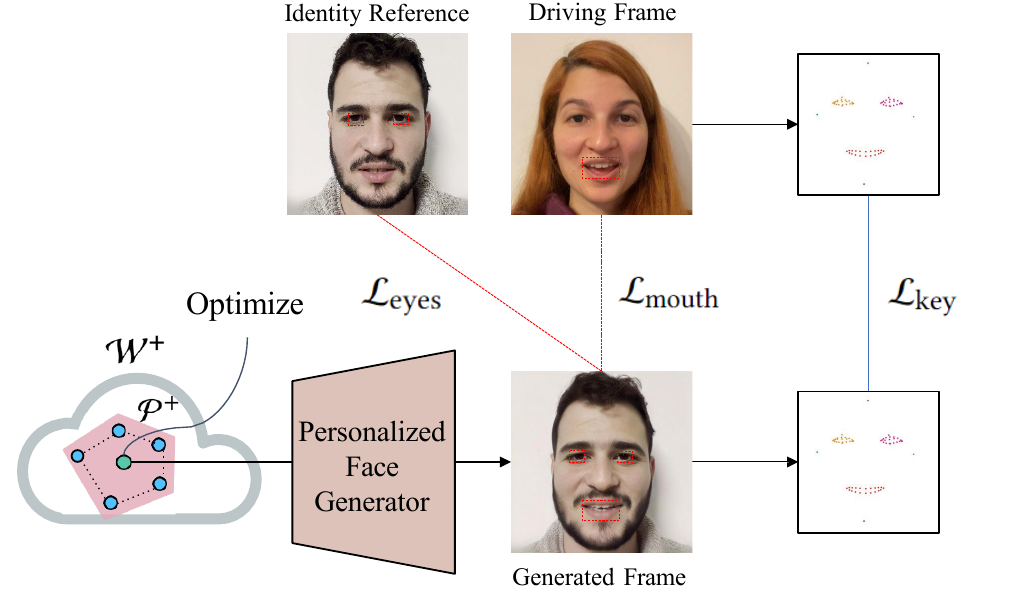}
    \caption{
    Illustration of our latent optimization method for generating a reenacted frame.
    We operate within the personalized latent space \pplus of the generator, ensuring identity preservation and reducing the task to reproducing pose and expression.
    To this end, we employ three losses to compare the generated and driving frames.
    First, we penalize differences in a subset of keypoints~\cite{kartynnik2019real}, which are separately normalized (indicated by color) for each facial part to have zero mean.
    The subset of normalized keypoints still conveys pose and expression but is more robust to geometry.
    Second, we apply an $L_2$ reconstruction loss on pixels in the mouth cavity between driving and generated frame for expression fidelity.
    Third, we apply an $L_2$ reconstruction loss on pixels in the irises areas between identity reference, which is generated from the center of \pplus, and generated frame for color preservation.
    }
    
    \label{fig:optimization}
\end{figure}

Identity preservation, the paramount challenge of reenactment, is guaranteed from using the personalized generator $G$.
Therefore, the task is reduced to accurately reproducing head poses and facial expressions of the driving frames.
To this end, we employ \emph{latent optimization} on each driving frame, directly optimizing a latent code such that the image generated from it would be suitable.
An overview of the process if illustrated in~\cref{fig:optimization}.

\subsubsection{Latent Space}
MyStyle's point-wise training (\cref{eq:mystyle_training}) affects only certain regions of the latent space, making them preserve the individual's identity~\cite{nitzan2022mystyle}.
These regions are modeled well by a slightly dilated convex hull defined by the anchors $\{w_i\}_{i=1}^N$.
To ensure identity preservation, \citet{nitzan2022mystyle} used the convex hull as the generator's latent space, instead of the latent spaces commonly used with StyleGAN, such as \w and \wplus \cite{abdal2019image2stylegan}.
We similarly adopt and make use of the personalized subspace, dubbed \pplus, for latent optimization.
A thorough and formal introduction of the latent space is provided in ~\cref{sec:supp_optimization}.

\subsubsection{Loss Function}

To guide optimization towards a latent code that replicates the desired head pose and facial expression, we use $3$D facial keypoints. 
Some aspects of expression, like the mouth cavity, are overlooked by keypoints. Therefore, we also employ a pixel-reconstruction loss directly in specific regions, as will be soon discussed. 

\myparagraph{Keypoints Loss.}
Indicating the locations and shapes of facial parts, facial keypoints hold rich information regarding head pose and facial expression.
However, keypoints are also affected by the unique appearance of each individual~\cite{nixon1985eye}.
Therefore, forcing a target identity to replicate keypoints of another individual leads to unsatisfactory results, and restricted previous works to only support self-reenactment~\cite{zakharov2019few}.
We propose an alternative guidance, derived from keypoints, that similarly indicates pose and expression but is significantly more robust to identity.

\begin{figure}
    \centering
    \begin{subfigure}{0.5\linewidth}
         \centering
         \includegraphics[width=\linewidth]{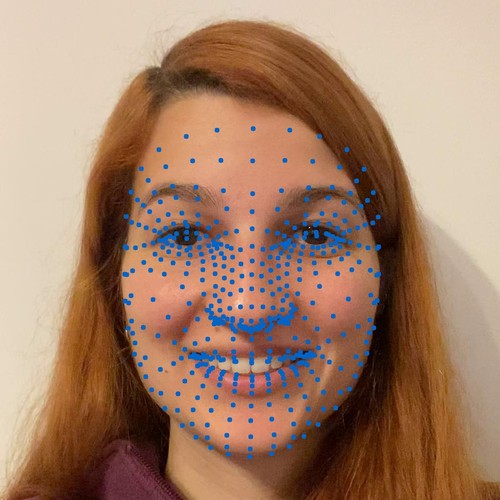}
         \label{fig:keypoints_all}
    \end{subfigure}%
    ~
    \begin{subfigure}{0.5\linewidth}
         \centering
         \includegraphics[width=\linewidth]{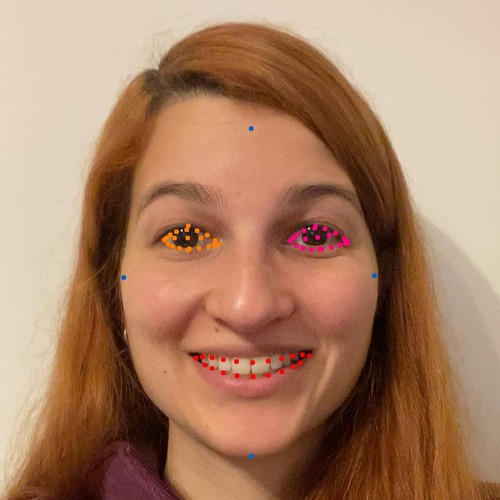}
         \label{fig:keypoints_used}
    \end{subfigure}
    \vspace{-8mm}
        \caption{
    Comparison of all keypoints inferred by MediaPipe~\cite{kartynnik2019real} (left) with our reduced subset (right).
    The subset chosen conveys pose and expression while being significantly less sensitive to identity.
    Colors other than blue indicate the grouping according to facial parts.
    The keypoints of each facial part are normalized separately to have zero mean.
    The blue landmarks are used with their absolute, non-normalized, coordinates.
    }
    \vspace{-3mm}
    \label{fig:keypoint_subset}
\end{figure}

Specifically, we start from a set of $468$ $3$D keypoints estimated using off-the-shelf method, MediaPipe~\cite{kartynnik2019real}, and make several modifications, illustrated in~\cref{fig:keypoint_subset}.

We first consider only a subset of all keypoints.
Most keypoints are essential for modeling face geometry, but carry very little semantic meaning.
Compare, for example, keypoints on the cheeks or forehead with those on the eyes and mouth.
The eyes and mouth are perceptually more significant and receive special attention in keypoint estimation~\cite{grishchenko2020attention} and reenactment~\cite{zakharov2019few}.
Even within the mouth, keypoints indicating whether the mouth is open or closed are crucial, but those that indicate the shape of the lips are person-specific and thus irrelevant.
Accordingly, we use a subset of keypoints that constitutes only of -- inner parts of both lips, eyelids and irises.

One can still deduce the distances between the eyes and mouth from our subset of keypoints.
This information is again person-specific and was even used directly in early face recognition methods~\cite{nixon1985eye}.
We overcome this problem by separately normalizing keypoints of the \emph{same} facial part to have zero mean. 
Since all keypoints are now relative, they cannot be used to deduce pose.
We therefore bring back four individual keypoints, indicating the frame of the head, which are used with their original absolute image coordinates.
In total, we use $66$ keypoints -- $16$ for each eye, $5$ for each iris, $20$ for the mouth and $4$ for the frame of the head.

Finally, the keypoint-based loss function is the sum of distances between keypoints of the driving and generated images.
Formally,
\begin{equation}
    \mathcal{L}_{\text{key}} = \sum_{i=1}^{66} \; \norm{\text{keypoints}(I_d)_i - \text{keypoints}(I_g)_i}_2,
\end{equation}
where \textit{keypoints()} is the operation extracting our subset of keypoints from the driving and generated images $I_d, I_g$, respectively. 

\myparagraph{Pixel Losses.}
Facial keypoints do not model some aspects of expression, most notably -- the mouth cavity, and solely relying on them hinders the fidelity of results~\cite{wang2023styleavatar,zakharov2019few,kim2018deep,zheng2022avatar}.
For example, even with a perfectly transferred lips motion, teeth might inappropriately appear or disappear from generated frames.
To mitigate this issue, we impose a loss that encourages actual pixel reconstruction between the mouth region of the driving and generated frames.
Formally, this loss reads as
\begin{equation}
\begin{split}
    & \mathcal{L}_{\text{mouth}} =
    \norm{ \text{mouth}(I_d) - \text{mouth}(I_g) }_2,
\end{split}
\end{equation}
where \textit{mouth} is the operation of cropping and resizing the mouth region, as defined by  keypoints, from the driving and generated images $I_d, I_g$ (see ~\cref{sec:supp_imp_details} for details).

At first, this loss might appear counter-productive as it encourages copying regions from the driving individual to the target individual -- harming identity preservation.
However, this is not a concern in our case, since using a personalized generator guarantees identity preservation.
The added pixel loss can only guide the optimization towards the ``nearest'' \textit{possible} solution, which is necessarily identity preserving.

We find that another region that requires special attention is that of the irises.
Aiming for an expressive reenactment setting, the irises should be able to convey any expression, such as looking right or down.
However, StyleGAN perpetuates biases inherited from FFHQ, where irises are almost always looking towards the camera.
We find the personalized generator is able to reconstruct the ``iris expressions'' seen in the self-scan but struggles to generalized beyond them.

In practice, the keypoint-driven optimization is able to reproduce the iris at the right location, at the price of its color -- often producing an off-white iris.
We find it possible to recover the correct eye color by adding a dedicated pixel reconstruction loss on the irises. 
Formally, the loss reads as
\begin{equation}
\begin{split}
    & \mathcal{L}_{\text{eyes}} =
    \norm{ \text{left-eye}(I_c) - \text{left-eye}(I_g) }_2 + \\
    & \norm{ \text{right-eye}(I_c) - \text{right-eye}(I_g) }_2,
\end{split}
\end{equation}
where \textit{left-eye and right-eye} are the operations of cropping and resizing these regions from (details in ~\cref{sec:supp_imp_details}) the generated image $I_g$ and an image serving as reference for the identity $I_c$ .
The reference image $I_c$ is generated from the center of the personalized subspace.

\myparagraph{Overall Loss.}

The overall reenactment loss reads as
\begin{equation}
\label{eq:opt_loss}
    \mathcal{L}_{\text{full}} = \mathcal{L}_{\text{key}} + \lambda_{\text{mouth}} \mathcal{L}_{\text{mouth}} + \lambda_{\text{eyes}} \mathcal{L}_{\text{eyes}} + \lambda_{\text{reg}}\mathcal{L}_{\text{reg}},
\end{equation}
where $\mathcal{L}_{\text{reg}}$ is a regularization term adopted from MyStyle that is partially responsible for optimization being constrained to the latent space \pplus, and is thoroughly explained in~\cref{sec:supp_optimization}.

\subsubsection{Temporal Consistency.}
Operating on each frame individually oftentimes leads to temporally incoherent results.
We follow the approach proposed by~\citet{tzaban2022stitch}, where temporal coherency is inherited from the input video by using smooth operations on individual frames.
We take several independent steps.

Similar to the training set, driving frames must also be aligned.
The crop and alignment performed are determined using facial keypoints~\cite{karras2019style}, which we smooth with a Gaussian low-pass filter.
Next, each optimization process is initialized with the result of the previous frame, with small additive Gaussian noise.
Adding noise is necessary to allow optimization escape the local minima of using the previous latent code again.
We find that this initialization is crucial for biasing the optimization to consistently converge to nearby latent codes, minimizing unnecessary perturbation.
We finally apply another Gaussian low-pass filter, this time on the produced latent codes themselves, making the transition between frames even smoother.

\section{Experiments}
In the following experiments we use $8$ self-scans of individuals from various genders, ages and ethnicities.
A sample of our results is displayed in~\cref{fig:our_subjects}, and additional videos are available in the supplemental.
We start by comparing our method to state-of-the-art facial reenactment methods, qualitatively and quantitatively in~\cref{subsec:comparison}.
In~\cref{subsec:edit}, we demonstrate that by performing reenactment through a semantic latent space, we can intuitively apply semantic editing and stylization in post-process.
We then evaluate the design choices of our method through an ablation study in~\cref{subsec:ablation}, and the generator's ability to generalize beyond the pretraining data and even the self-scan data by novelly composing expressions in~\cref{subsec:generalization}.

\subsection{Comparison}
\label{subsec:comparison}

\myparagraph{Baselines.}
We first compare our method with three state-of-the-art methods.
We choose works representative of several of the main branches of facial reenactment literature, as described below.
Neural Head Avatars (NHA)~\cite{grassal2022neural} receives as input a single video and explicitly models $3$D
mesh surface. Reenactment is performed by transferring parameters between $3$D morphable models~\cite{li2017learning}. We note that NHA's video is ``proffesionally'' taken  -- \ie taken in front of a green screen, with even lighting, and is also $\times3$ longer than ours.
Latent Image Animator (LIA)~\cite{wang2022latent} is a \textit{one-shot} method, taking in a single image as input, and warping it to match driving frames. The warp is determined according to latent controls learned by a 2D generative model trained for that purpose. 
DaGAN~\cite{hong2022depth} is a one-shot method, using a dedicated $2$D image-to-image translation module, translating each driving frame to a reenactment frame.

\myparagraph{Comparison Protocol.}
\begin{table}[t]
    \centering
    \setlength{\tabcolsep}{1.5pt}
    \caption{
    Comparison to competing methods, in terms of identity preservation (IDP), identity consistency (IDC), average expression distance (AED), average pose distance (APD), and  NIQE~\cite{mittal2012making}.
    }
    \begin{tabular}{lccccc}
         \toprule
         &
         ID$(\uparrow)$ &
         APD{\tiny $\times100$} $(\downarrow)$ &
         AED{\tiny $\times10$} $(\downarrow)$ &
         NIQE $(\downarrow)$ \\
         \midrule
         NHA & 0.80 & 0.24 & 2.37 & 19.34  \\
         LIA & 0.71 & \textbf{0.13} & 1.59 & 45.52   \\
         DaGAN & 0.73 & 0.26 & 1.77 & 34.88 \\
         Ours & \textbf{0.89} & 0.16 & \textbf{1.52} & \textbf{13.22} \\
         \bottomrule
    \end{tabular}
    \label{tab:quant}
\end{table}

Here, we use $4$ of our self-scans videos.
We use each video separately as a self-scan, and also to drive the other self-scans.
We use random $5\text{-}10$ seconds sequences from the driving videos and produce a total of \work{$10$} results.
We note that one-shot methods are given the first frontal-pose, neutral-expression frame in the video of the target individual, and video-based methods use entire video. NHA also requires an indication of one such frame.

We use multiple quantitative metrics to evaluate the performance of reenactment.
First, We measure identity preservation (ID) with respect to the training set.
To this end, we follow a protocol similar to MyStyle's, where for each reenacted frame we compute its average similarity to the top-$5$ most similar images in the training set and average across all frames.
We use top-$5$ images (instead of MyStyle's top-$1$), to prevent one-shot methods from being trivially similar to their reference image.
Similarity is measured in terms of cosine similarity of features extracted from a face recognition network~\cite{deng2019arcface}.

Next, we measure pose and expression preservation from the driving video.
Following previous works~\cite{sun2022next3d,lin20223d}, we measure Average Pose Distance (APD) and Average Expression Distance (AED). 
Both are defined as the mean square error of deep features extracted by a $3$D face reconstruction model~\cite{deng2019accurate}.
Last, we evaluate the frame's quality.
Due to the few-shot nature of the problem, popular metrics such as FID are unsuitable.
We instead use NIQE~\cite{mittal2012making}, a metric designed to evaluate image quality without reference.

We report the metrics in~\cref{tab:quant}, a set of qualitative results in~\cref{fig:qual}, and videos in the supplemental.
As can be seen, our method outperform all existing methods on identity and expression preservation, and quality, and performs competitively with the leader LIA on pose preservation. 
Specifically, by comparing generated images to reference self-scan frames, one can observe that our method excels at preserving identity.

\subsection{Edits and Stylization in Post-Process}
\label{subsec:edit}

\begin{figure}[h]
    \centering
     \includegraphics[width=\linewidth]{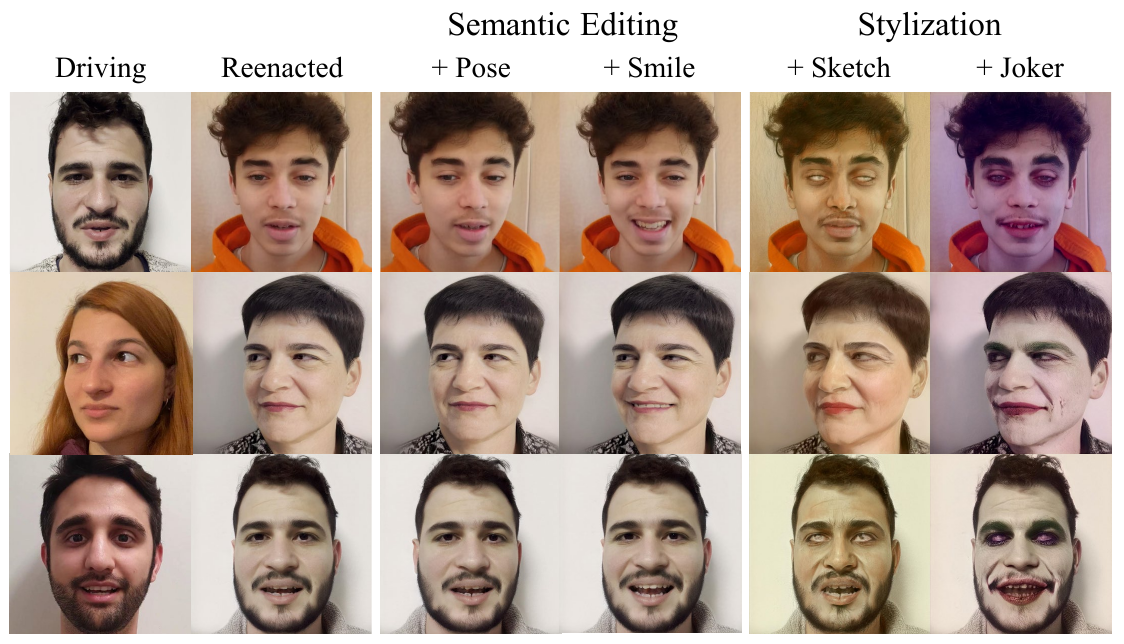}
    \caption{
    Semantic editing and stylization of the reenacted video performed in post-process.
    Before being generated, the reenacted video is represented as a sequence of codes in StyleGAN's latent space.
    The reenacted video can thus be semantically edited by  shifting these latent codes in semantic directions~\cite{shen2020interfacegan} before feeding them to the generator, or stylized by feeding them to a generator that was fine-tuned for a specific style~\cite{gal2021stylegan}.
    }
    \label{fig:edits}
\end{figure}

Our reenactment method is built on top of an incredibly semantic image generator -- StyleGAN.
We can therefore leverage the myriad of applications and methods designed for StyleGAN, adding new capabilities on top of reenactment.
In this section we demonstrate two such capabilities, performed in post-process -- semantic editing and stylization. 
Results for both applications are presented in~\cref{fig:edits}. 

\myparagraph{Semantic Editing.}
In StyleGAN's latent space there exist vectors or ``directions'' that when traversed upon, affect the generated image in a semantically consistent manner. For example, one such vector would consistently and gradually add a smile.
The result of our latent optimization method is a set of latent codes, that turn into the reenacted video once passed through the generator.
Here, instead of generating frames from these latent codes, we first shift them a constant sized step in the direction of a specific semantic vector.
Passing these shifted latent codes through the generator results in a reenacted video that was semantically additionally edited.
We use three semantic vectors that are identified using InterFaceGAN~\cite{shen2020interfacegan} -- smile, pose, and age.

\myparagraph{Stylization.}
Several methods~\cite{gal2021stylegan,ojha2021few} have been proposed to fine-tune StyleGAN slightly so that it models different data domains.
These fine-tuned models are known to be \textit{aligned} with their parent models, \ie images generated by the two models, from the same latent code, correspond to each other~\cite{wu2021stylealign}.
Here, we further fine-tune our personalized models with StyleGAN-NADA~\cite{gal2021stylegan}, a text-driven training method.
We fine-tune with two prompts -- ``the Joker'' and ``a sketch''.
The latent optimization still takes place in the personalized generator, but the resulting latent codes are passed through the generator tuned with StyleGAN-NADA to produce the final video frames.
The result is a reenacted video which depicts the target individual but they are stylized as the Joker or a sketch. 

As can be seen in~\cref{fig:edits}, the reenacted frames are successfully edited and stylized. 
This native post-process capability greatly extends the potential applications of our method and further encourages the use of semantic generative models as backbones for reenactment.

\subsection{Ablation Studies}
\label{subsec:ablation}

\begin{figure}[th]
    \centering
     \includegraphics[width=\linewidth]{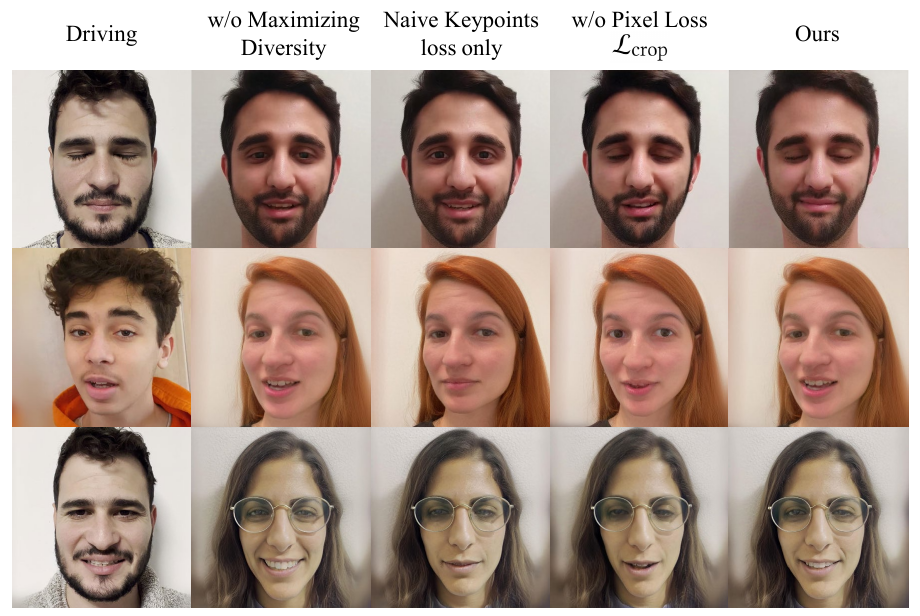}
    \caption{
    Ablation study.
    Evenly spaced self-scan frames miss uncommon expressions, like closing eyes, prohibiting from reproducing them (\eg, first row).
    The naive keypoint-driven reenactment method does not generalize between different facial geometries and completely fails transferring expressions.
    Last, adding the pixel loss $\mathcal{L}_{\text{crop}}$, generally improves the expression preservation efficacy and specifically encourages accurate appearance of teeth in the mouth cavity.
    }
    \vspace{-3mm}
    \label{fig:ablations}
\end{figure}

We next evaluate the contribution of individual aspects of our method through an ablation study.
A sample of results is provided in~\cref{fig:ablations}.

The personalized generator is trained on a subset of self-scan frames that maximizes diversity.
Here, we instead train personalized models on frames that are evenly spaced across the self-scan, using the same number of frames $N=250$.
We find that these models fail to reproduce relatively uncommon expressions, such as closing the eyes.
The reason becomes immediately clear when looking at the training set -- these expressions, purely by chance, are entirely not present.

We move on to consider the latent optimization process. 
First, we use the naive optimization loss for keypoint-driven reenactment -- penalizing differences in absolute keypoint coordinates.
We find that all expressions are poorly reproduced. This is expected, since these keypoint representations are entangled with facial geometry and do not trivially correspond across individuals.
Next, we reinstate our keypoint loss $\mathcal{L}_\text{key}$, but still without any pixel loss.
While pose and expression are better preserved, differences from the driving frames are still clear, most notably in the shape of the lips and the appropriate appearance of teeth in the mouth cavity.
Our final method, adding the pixel loss $\mathcal{L}_\text{crop}$ immediately mitigates these issues.

\subsection{Studying the Generator's Expressiveness}
\label{subsec:generalization}

\begin{figure}[h]
    \centering
     \includegraphics[width=\linewidth]{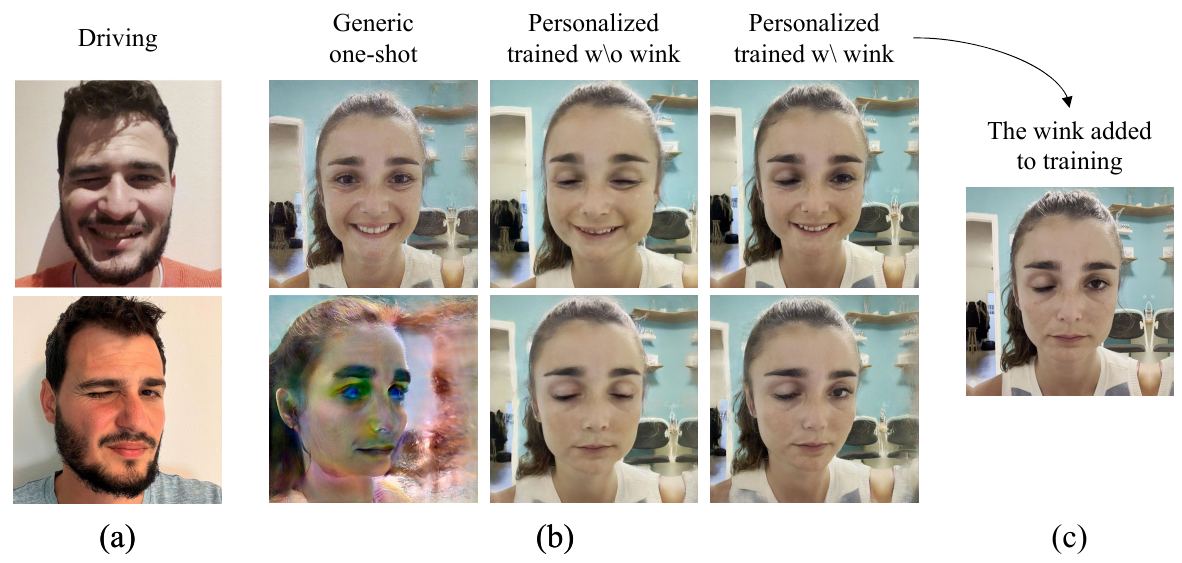}
    \caption{
    Evaluating different generator's ability to produce a wink.
    We reenact a target individual with two winking driving frames, seen in (a), through different models (b). 
    The generic StyleGAN model, used in a one-shot manner, fails to produce winking faces and severely degrades the generated images in the process.
    A personalized model whose training set does not include a wink also fails to a wink. 
    Nevertheless, the optimization being confined to the generator's personalized subspace assures that the produced images are not degraded and faithfully preserve the individual's appearance.
    Last, adding a single instance of winking (c) to the training set, allows the personalized generator to produce winking images.
    Note that the generated images compose a wink with a smile and a head pose that were not seen in training.
    }
    \label{fig:convex_expressiveness}
\end{figure}

Our reenactment through latent optimization could be considered a retrieval of the most suitable frame from the personalized generator.
It is thus intriguing to ask -- \textit{what can and cannot it generate?}
For this study, we use the relatively unusual expression of winking.
Results are displayed in~\cref{fig:convex_expressiveness}.

We first look at the generic StyleGAN model trained on FFHQ~\cite{karras2019style}.
We apply our latent optimization method, using an inverted image~\cite{roich2021pivotal} as initialization and a winking driving frame as guidance.
This approach produces results that are not winking and are severely degraded.

We now capture a self-scan that is scripted to intentionally include a single instances of winking, with a neutral faces expression and frontal pose.
From the self-scan we create two training set, where one is created normally and the from the other we omit the few winking frames.
We train a personalized model with each of these training sets.
Applying our reenactment method with winking driving frames to the two models, we find that only the model that has seen a wink in training is able to generate winking frames.
We note that the generated winking frames are varied, composing a wink with other expressions and poses that were not seen in the single winking training instance.

We deduce two important conclusions.
First, including unique expressions, and possibly other information, extends the model's expressive power beyond what is provided by pre-training.
On the flip side, an expression that is absent from the self-scan may not be produced.
Second, the model can compose expressions and poses that were never seen together in training. 
This allows self-scans to be casually captured, not requiring a strict script and an exhaustive coverage of all setting combinations.

\begin{figure*}
    \centering
     \includegraphics[width=0.9\linewidth]{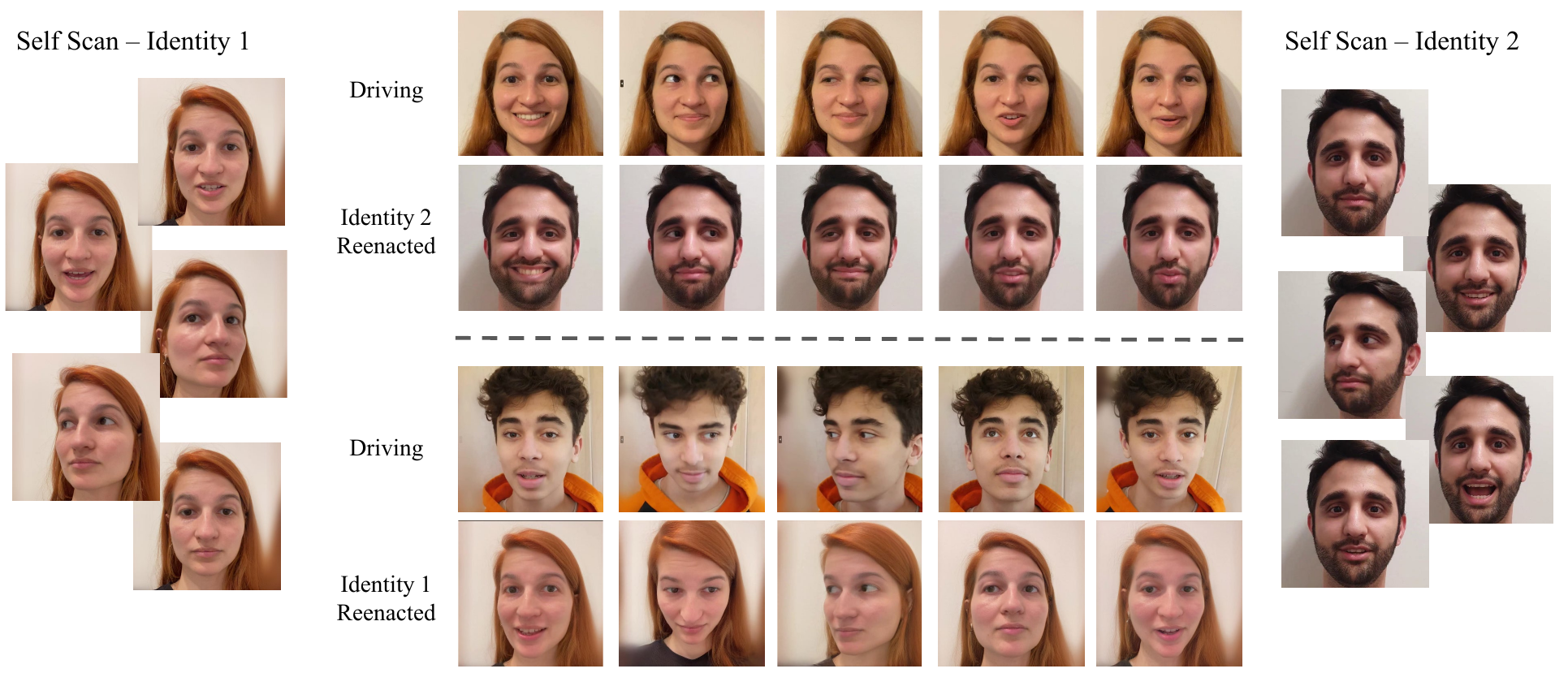}
    \caption{
    A sample of qualitative reenactment results produced by our method (center).
    Frames from the relevant self-scan are provided on both sides.
    Our method successfully reproduces poses and expression from the driving video while faithfully preserving the target identity.
    We recommend zooming in and carefully evaluating identity preservation by carefully comparing generated and self-scan frames.
    }
    \label{fig:our_subjects}
\end{figure*}

\begin{figure*}
    \centering
     \includegraphics[width=0.9\linewidth]{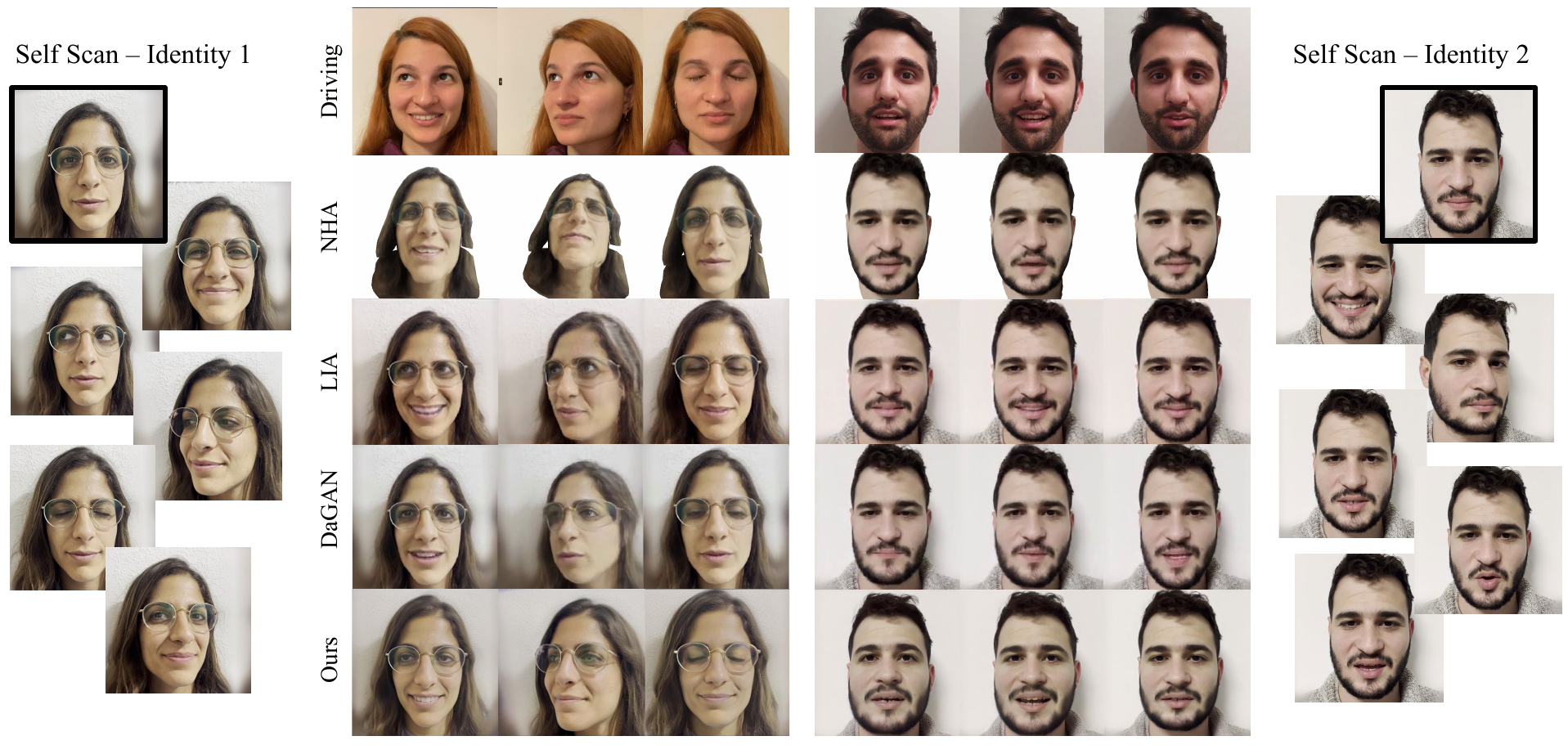}
     \caption{
    Comparing ours and state-of-the-art reenactment methods -- NHA~\cite{grassal2022neural}, LIA~\cite{wang2022latent} and DaGAN~\cite{hong2022depth}.
    Frames from the self-scans of the target identity are provided on both sides.
    One-shot methods -- LIA and DaGAN -- take as input only the top image, marked with a black outline.
    As can be seen, our results are at least on par at transferring pose and expression and  significantly better at realism and preserving identity.
    We recommend zooming in and carefully evaluating identity preservation by carefully comparing generated and self-scan frames.
    }
    \label{fig:qual}
\end{figure*}

\section{Discussion and Conclusion}

\myparagraph{Limitations.}
We have demonstrated that our method produces state-of-the-art results.
Nevertheless, our method has a few limitations.
First, as discussed in~\cref{subsec:generalization,fig:convex_expressiveness}, our generator is bounded by the variations appearing in the self-scan.
Driving frames that depict unseen behavior, mostly expressions, would produce low fidelity results.
Second, the optimization process performed per frame is time-consuming, requiring up to a minute for a single frame.

\myparagraph{Conclusion.}
We have proposed a novel facial reenactment method leveraging a personalized image generator, trained on a casual self-scan of a single individual.
Identity preservation is guaranteed by the generator being personalized, and thus we have reduced reenactment to reproducing poses and expressions, solved using a latent optimization process.
Our experiments and analysis demonstrated that our method outperforms prior work in quality and fidelity, specifically excelling at identity preservation.

This study highlights the potential of $2$D generative models for reenactment beyond the one/few-shot setting, effectively bridging a gap between the $2$D and $3$D streams of works.
We hope it will inspire others to pursue this direction further.
A natural first step would be to develop an encoder to replace latent optimization, substantially accelerating the process.

\section*{Acknowledgments}
We are grateful to Or Patashnik for insightful discussions throughout this work.
We also thank Or Patashnik and Omer Tov for proofreading the draft.
We finally thank Michal Elazary, Yogev Nitzan, Daniel Garibi, Zvia Elazary, Inbal Raz and Noa Rubin for providing us with their self-scans.
This research was supported in part by the Israel Science Foundation (grants no. 2492/20 and 3441/21).

{
\small
\bibliographystyle{ACM-Reference-Format}
\bibliography{main}
}

\appendix
\section{Supplemental Overview}

In \cref{sec:supp_optimization} we elaborate on details related to the latent optimization process that were kept out of the main paper in sake of clarity.
Then, \cref{sec:supp_imp_details} provides technical implementation details.

\section{Latent Space for Optimization}
\label{sec:supp_optimization}
At the core of our method is a latent optimization process in the latent space of a personalized generator, trained with MyStyle~\cite{nitzan2022mystyle}.
To ensure identity preservation, MyStyle use a novel latent space for all its downstream application.
In this work, we adopt this latent space, without any modifications.
This section thoroughly describes how we use that latent space, partially repeating discussions appearing in MyStyle for the sake of completeness.

\subsection{Native Latent Space, \p}
\citet{nitzan2022mystyle} observed that only a certain region of the generator's latent space has been personalized as result of training (see~\cref{subsec:training}).
To capture this region, they propose using a slightly ``dilated'' convex hull spanned by the anchors $\{w_i\}$.
Formally, they define
\begin{equation}
\label{eq:p_b}
    \mathcal{P}_{\beta} 
    = \{ \sum_i \alpha_i w_i \vert \, \sum_i \alpha_i = 1, \forall i: \alpha_i \geq -\beta \},
\end{equation}
where $\beta$ is a positive, typically small, number defining the amount of dilation.
Note that \pbeta is a subspace of the generators \w latent space.
The hyperparameter $\beta$ is usually omitted from notation, and we instead denote the space as \p.
With respect to \p, they choose to represent latent codes in Barycentric coordinates~\cite{floater2015generalized}, \ie, using the convex coefficients $\{\alpha_i\}$.
Optimization is thus technically performed on a vector of these coefficients -- $\vect{\alpha}$.
Changing back to standard coordinates is trivially done by $\sum_i \alpha_i w_i$.

Several measures are taken to ensure that the optimization process remains within the bounds of \pbeta.
First, the vector $\vect{\alpha}$ is passed through a shifted softplus function: 
\begin{equation}
\label{eq:softplus}
 \tilde{\vect{\alpha}} = \frac{1}{s}log(1+e ^ {s (\vect{\alpha} + \beta)}) - \beta,
\end{equation}
where all operations are element-wise, and $s$ is a ``sharepness'' hyperparameter we set to $s=100$.
This operation yields a new vector $\tilde{\vect{\alpha}}$ whose minimal value is greater than $-\beta$.

Next, a regularization term is added to the optimization loss to encourage the sum of coefficients to be $1$.
Formally,
\begin{equation}
\label{eq:sum_reg}
    \mathcal{L}_{\text{sum}}(\tilde{\vect{\alpha}}) = (\sum_i \tilde{\alpha}_i - 1 ) ^2.
\end{equation}
Finally, the optimization result $\vect{\alpha}^*$ is explicitly normalized to have a sum of $1$.

\subsection{Extended Latent Space, \pplus}

First proposed by~\citet{abdal2019image2stylegan}, an incredibly popular approach to increase StyleGAN's expressivity is using an extended latent space \wplus~\cite{richardson2021encoding,menon2020pulse,luo2020time}.
Concisely, this latent space is composed by using a possibly different latent code in each layer of the generator.
With its greater expressivity, it was also shown that \wplus provides a less faithful prior.
This has led previous works to regularize the extent to which it is exploited, \eg, by encouraging latent code in different layers to still be similar under $L_2$ norm~\cite{tov2021designing}.

\citet{nitzan2022mystyle} followed these ideas directly and imported them to \p.
First, \p is extended to \pplus across generator's layers by allowing each layer to add an offset: $\vect{\alpha}_{\beta}^{+} = (\vect{\alpha}_{\beta} + \vect{\Delta}_0, ..., \vect{\alpha}_{\beta} + \vect{\Delta}_N)$
Second, a regularization term is added to the optimization process and prevent severe deviation across layers.
Formally,
\begin{equation}
\label{eq:delta_reg}
    \mathcal{L}_{\text{delta}}(\Delta) = \sum_{i} \norm{\vect{\Delta_i}}_{2}.
\end{equation}

\subsection{Final Optimization loss}
The two regularization terms described in~\cref{eq:delta_reg,eq:sum_reg} were jointly referred to in the main paper's~\cref{eq:opt_loss} as regularization terms assuring the optimization is constrained to the personalized latent space.
The full optimization loss is thus given by
\begin{equation}
\label{eq:supp_opt_loss}
    \mathcal{L}_{\text{full}} = \mathcal{L}_{\text{key}} + \lambda_{\text{mouth}} \mathcal{L}_{\text{mouth}} + \lambda_{\text{eyes}} \mathcal{L}_{\text{eyes}} + \lambda_{\text{delta}}\mathcal{L}_{\text{delta}}+\lambda_{\text{sum}}\mathcal{L}_{\text{sum}}.
\end{equation}

\begin{algorithm}
\caption{Greedy algorithm to maximize subset's diversity~\cite{ravi1994heuristic}.}
\label{alg:subset}

\KwData{Set $S$, distance function $d$, subset size $N$.}
Let $s_i, s_j$ be the elements of set $S$ who have the greatest distance $d$.
$T \gets \{s_i, s_j\}$

\While{$|T| < N$}
{
    $s' \gets \underset{s \in S \setminus T}{\argmax} \ \underset{t \in T}{\min} \ d(s, t)$\;
    $T \gets T \cup \{s'\}$\;
}
Return $T$.
\end{algorithm}

\section{Additional Technical Details}
\label{sec:supp_imp_details}

\subsection{Self-Scan Frame selection}
The formal description of the frame selection algorithm is provided in~\cref{alg:subset}.

\subsection{Crops of the Mouth and Eyes}
Our optimization's pixel loss $\mathcal{L}_{\text{crop}}$ is calculated on crops of the mouth and eyes that are automatically extracted from the generated and driving frames.
We next detail how these crops are obtained.
We crop the minimal rectangles that bounds the keypoints from our subset for each region of interest -- the outer parts of the lips and the irises.
We then take center crops of the rectangles from the larger of the two rectangles, on each axis, to have the rectangles at the same size.

\subsection{Implementation details}
We next specify all hyperparameters.
For the latent space, we use $\beta=0.02$ in all experiments.
The weights of the different loss terms $\lambda_{\text{crop}}=1, \lambda_{\text{delta}}=10, \lambda_{\text{sum}}=10$.
The additive Gaussian noise for the last frame result has $\sigma=0.05$.
The Gaussian low-pass filter on the produced latent codes has $\sigma=2$.
The number of self-scan frames in all experiments $N=250$.

Training takes roughly $3.5$ hours on a single $V100$, and optimization requires up to a minute for a single frame.

\end{document}